\documentclass[]{hdsr}

\usepackage{graphicx}
\usepackage{hyperref}
\usepackage[capitalize]{cleveref}  
\crefname{section}{Sec.}{Secs.}
\crefname{appendix}{App.}{Apps.}
\crefname{figure}{Tab.}{Tabs.}
\crefname{figure}{Fig.}{Figs.}
\usepackage{color}
\usepackage{subcaption}
\usepackage{wrapfig}

\DeclareLanguageMapping{english}{english-apa}

\graphicspath{{figs/}}

\begin{document} 

\newgeometry{bottom=1.5in}

\volumeheader{0}{0}{00.000}

\begin{center}

  \title{DeepMiner: Discovering Interpretable Representations for Mammogram Classification and Explanation}
  \maketitle

  \thispagestyle{empty}

  \begin{tabular}{cc}
    Jimmy Wu\upstairs{\affilone,*},~Bolei Zhou\upstairs{\affiltwo},~Diondra Peck\upstairs{\affilthree},~Scott Hsieh\upstairs{\affilfour},~Vandana Dialani, MD\upstairs{\affilfive},\\
    Lester Mackey\upstairs{\affilsix},~and Genevieve Patterson\upstairs{\affilseven,*}
   \\[0.25ex]
   {\small \upstairs{\affilone} Department of Computer Science, Princeton University, Princeton, USA} \\
   {\small \upstairs{\affiltwo} The Chinese University of Hong Kong, Hong Kong, China} \\
   {\small \upstairs{\affilthree} Microsoft, Redmond, USA} \\
   {\small \upstairs{\affilfour} Department of Radiological Sciences, UCLA, Los Angeles, USA} \\
   {\small \upstairs{\affilfive} Beth Israel Deaconess Medical Center, Cambridge, USA} \\
   {\small \upstairs{\affilsix} Microsoft Research New England, Cambridge, USA} \\
   {\small \upstairs{\affilseven} VSCO, Oakland, USA} 
  \end{tabular}
  
  \emails{
    \upstairs{*}jw60@cs.princeton.edu, genevieve@vsco.co
    }
  \vspace*{0.4in}

\begin{abstract}
We propose DeepMiner, a framework to discover interpretable representations in deep neural networks and to build explanations for medical predictions. By probing convolutional neural networks (CNNs) trained to classify cancer in mammograms, we show that many individual units in the final convolutional layer of a CNN respond strongly to diseased tissue concepts specified by the BI-RADS lexicon. After expert annotation of the interpretable units, our proposed method is able to generate explanations for CNN mammogram classification that are consistent with ground truth radiology reports on the Digital Database for Screening Mammography. We show that DeepMiner not only enables better understanding of the nuances of CNN classification decisions but also possibly discovers new visual knowledge relevant to medical diagnosis.
\end{abstract}
\end{center}

\vspace*{0.15in}
\hspace{10pt}
  
\copyrightnotice

\section*{Media Summary}
Deep learning algorithms are often criticized for producing uninterpretable, black-box results. This lack of interpretability is especially concerning for medical decisions. Here we introduce DeepMiner, a human-in-the-loop framework for interpreting the medical predictions of deep neural networks.
After training a neural network to detect malignancy in mammograms, DeepMiner identifies the visual phenomena most influential to the network's decision-making and calls upon human experts to label those phenomena (for example as ``calcified vessels,'' ``masses with smooth edges,'' or ``normal breast tissue''). These expert annotations enable DeepMiner to explain its mammogram classifications by summarizing the named phenomena driving each prediction.

\restoregeometry
\newgeometry{bottom=0.5in}

\section{Introduction}
Deep convolutional neural networks (CNNs) have made
great progress in visual recognition challenges such as object classification \citep{krizhevsky2012imagenet} and scene recognition \citep{zhou2014learning}, even reaching human-level image understanding in some cases \citep{he2015delving}. Recently, CNNs have been widely used in medical image understanding and diagnosis \citep{rajpurkar2017chexnet,esteva2017dermatologist,wang2017chestx}. However, with millions of model parameters, CNNs are often treated as `black-box' classifiers, depriving researchers of the opportunity to investigate what is learned inside the network and explain the predictions being made. Especially in the domain of automated medical diagnosis, it is crucial to have interpretable and explainable machine learning models. 

Several visualization methods have previously been proposed for investigating the internal representations of CNNs. For example, internal units of a CNN can be represented by reverse-mapping features to the input image regions that activate them most~\citep{zeiler2014visualizing} or by using backpropagation to identify the most salient regions of an image~\citep{Simonyan14a,DBLP:journals/corr/MahendranV14}. Our work is inspired by recent work that visualizes and annotates interpretable units of a CNN using Network Dissection~\citep{bau2017network}.

Meanwhile, recent work in automated diagnosis methods has shown promising progress towards interpreting models and explaining model predictions. \citet{wu2018expert} show that CNN internal units learn to detect medical concepts which match the vocabulary used by practicing radiologists. \citet{rajpurkar2017chexnet} and \citet{wang2017chestx} use the class activation map defined by \citet{zhou2016learning} to explain informative regions relevant to final predictions. \citet{Zhang_2017_CVPR} propose a hybrid CNN and LSTM (long short-term memory) network capable of diagnosing bladder pathology images and generating radiological reports if trained on sufficiently large image and diagnostic report datasets. However, their method requires training on full medical reports. In contrast, our approach can be used to discover informative visual phenomena spontaneously with only coarse training labels. \citet{jing2017automatic} successful created a visual and semantic network that directly generates long-form radiological reports for chest X-rays after training on a dataset of X-ray images and associated ground truth reports. However, even with these successes, many challenges remain. \citet{wu2018expert} only show that interpretable internal units are correlated with medical events without exploring ways to explain the final prediction. The heatmaps generated by \citet{rajpurkar2017chexnet,wang2017chestx} qualitatively identify important \emph{locations} in an image but fail to identify \textit{specific concepts}. \citet{jing2017automatic} train their models on large-scale medical report datasets; however, large text corpora associated with medical images are not easily available in other scenarios. Additionally, \citet{Zhang_2017_CVPR} acknowledge that their current classification model produces false alarms from which it cannot yet self-correct.

In this paper, we propose a general framework called \textit{DeepMiner} for discovering medical phenomena in coarsely labeled data and generating explanations for final predictions, with the help of a few human expert annotators. We apply our framework to mammogram classification, an already well-characterized domain, in order to provide confidence in  the capabilities of deep neural networks for discovery, classification, and explanation. 

To the best of our knowledge, our work is the first automated diagnosis CNN that can both discover discriminative visual phenomena for breast cancer classification and generate interpretable, radiologist-collaborative explanations for its decision-making. Our main contribution is two-fold: (1) we propose a human-in-the-loop framework to enable medical practitioners to explore the behavior of CNN models and annotate the visual phenomena discovered by the models, and (2) we leverage the internal representations of CNN models to explain their decision making, without the use of external large-scale report corpora.
Our data, results, and open source code replicating all experiments can be found at \url{https://github.com/jimmyyhwu/ddsm-visual-primitives}. A copy of our data and pretrained CNN weights is also available through the Harvard Dataverse \citep{dataset2021}.

\section{The DeepMiner Framework} 

\begin{figure}[h]
    \centering
        \includegraphics[width=\textwidth]{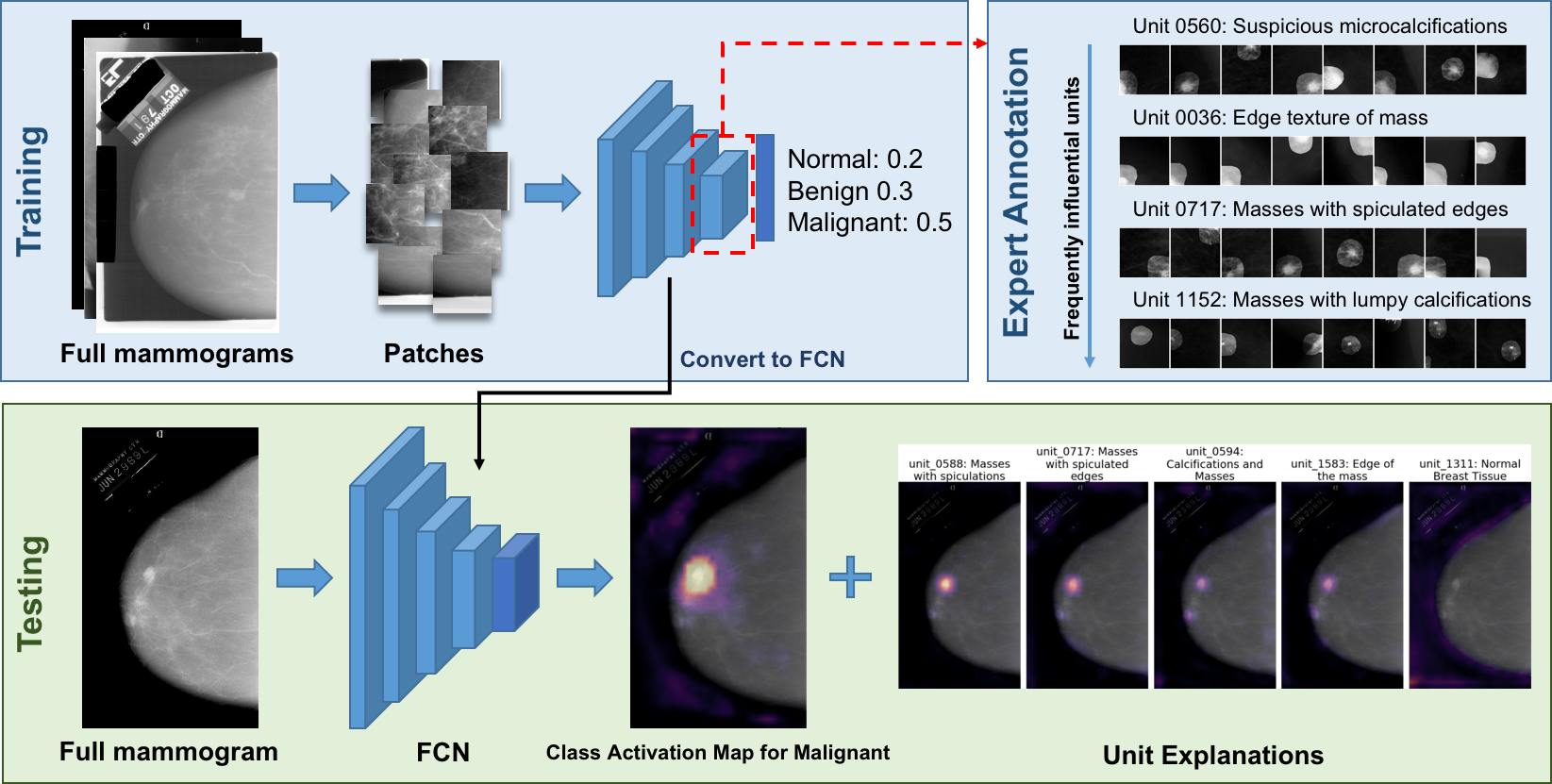}
	\caption{Illustration of the DeepMiner framework for mammogram classification and explanation.}
        \label{fig:deepminer}
\end{figure}

The DeepMiner framework consists of three phases, as illustrated in Fig. \ref{fig:deepminer}. In the first phase, we train standard neural networks for classification on patches cropped from full mammograms. Then, in the second phase, we invite human experts to annotate the top class-specific internal units of the trained networks. Finally, in the third phase, we use the trained network to generate explainable predictions by ranking the contributions of individual units to each prediction.

In this work, we select mammogram classification as the testing task for our DeepMiner framework. The classification task for the network is to correctly classify mammogram patches as normal (containing no findings of interest), benign (containing only non-cancerous findings), or malignant (containing cancerous findings). Our framework can be further generalized to other medical image classification tasks. Note that we refer to convolutional filters in our CNNs as `units', as opposed to `neurons', to avoid conflation with the biological entities.

\subsection{Dataset and Training} 

We choose ResNet-152 pretrained on ImageNet~\citep{He2015} as our reference network due to its outstanding past performance on object classification, image segmentation, and fine-grained localization across a variety of image domains \citep{he2017mask}.  
We replaced the final layer of the network with a 3-class classification layer and fine-tuned all network weights to classify mammogram patches as normal, benign, or malignant using data from the Digital Database for Screening Mammography (DDSM)~\citep{heath2000digital}. DDSM is a dataset compiled to facilitate research in computer-aided breast cancer screening. It consists of 2,620 cases, each including two images of each breast, a BI-RADS rating of 0-5 for cancer risk, a radiologist's subjective subtlety rating for each finding, and a BI-RADS keyword description of abnormalities. Labels include image-wide designations (e.g., malignant, benign, and normal) and pixel-wise segmentations of lesions~\citep{heath2000digital}.

In our experiments, we train our classifier on 80\% of the DDSM cases, reserve 10\% of the cases as a hold-out set to select the learning rate for training, and use the final 10\% of the images as a test set for evaluating CNN classification performance. 
We highlight that all images belonging to the same case are placed in the same dataset partition to mimic the standard use-case of evaluating a trained system on a previously unseen case.

To increase the number of training examples for fine-tuning, we extract smaller image patches from our mammograms in a sliding window fashion. The dimensions of each image patch are 25\% of the width of the original mammogram, and overlapping patches are extracted using a stride of 50\% of the patch width. 
We use a background texture classifier to discard any patch containing less than 50\% breast tissue.
We create three class labels (normal, benign, malignant) for each image patch based on (1) whether at least 30\% of the patch contains benign or malignant tissue and (2) whether at least 30\% of a benign or malignant finding is located in that patch.
These patch labels were determined automatically by calculating pixel overlap with the ground-truth lesion segmentation.

We fine-tune our reference network using the PyTorch \citep{NEURIPS2019_9015} implementation of stochastic gradient descent \citep{robbins1951stochastic,kiefer1952stochastic,bottou2018optimization}.  
The network was trained for 5 epochs with a median runtime of 85 minutes per epoch on a single Titan X Pascal GPU. 
We choose a batch size of $32$ to ensure that all images in the batch fit into GPU memory, select a learning rate of $0.0001$ to balance overfitting and underfitting as judged by differences in training set and hold-out set accuracies, and set the momentum and weight decay hyperparameters to the default values for ImageNet training in PyTorch ($0.9$ and $0.0001$, respectively).
The final test set performance of the trained network is presented in \cref{sec:res_class}, and we provide a brief introduction to CNNs and their training in \cref{sec:cnn-intro}.

\subsection{Human Annotation of Visual Primitives Used by CNNs}

We use our hold-out split of DDSM to create visualizations for units in the final convolutional layer of our fine-tuned ResNet-152. We choose the final layer since it is most likely to contain high-level semantic concepts due to the hierarchical structure of CNNs.

It would be infeasible to annotate all 2048 units in the last convolutional layer. Instead, we select a subset of the units deemed most frequently `influential' to classification decisions. Given a classification decision for an image, we define the influence of a unit towards that decision as the unit's maximum activation score on that image multiplied by the weight of that unit for a given output class in the final fully connected layer.

We selected the twenty most frequently influential units for each of the three classes and asked human experts to annotate the resulting 60 units. For the normal tissue class, if the twenty units we selected were annotated, those annotations would account for 59.27\% of the per-image top eight units over all of the hold-out set images. The corresponding amount for the benign class is 69.77\% and for the malignant class is 75.82\%. 

\begin{figure}[h]
    \centering 
    \includegraphics[width=\textwidth]{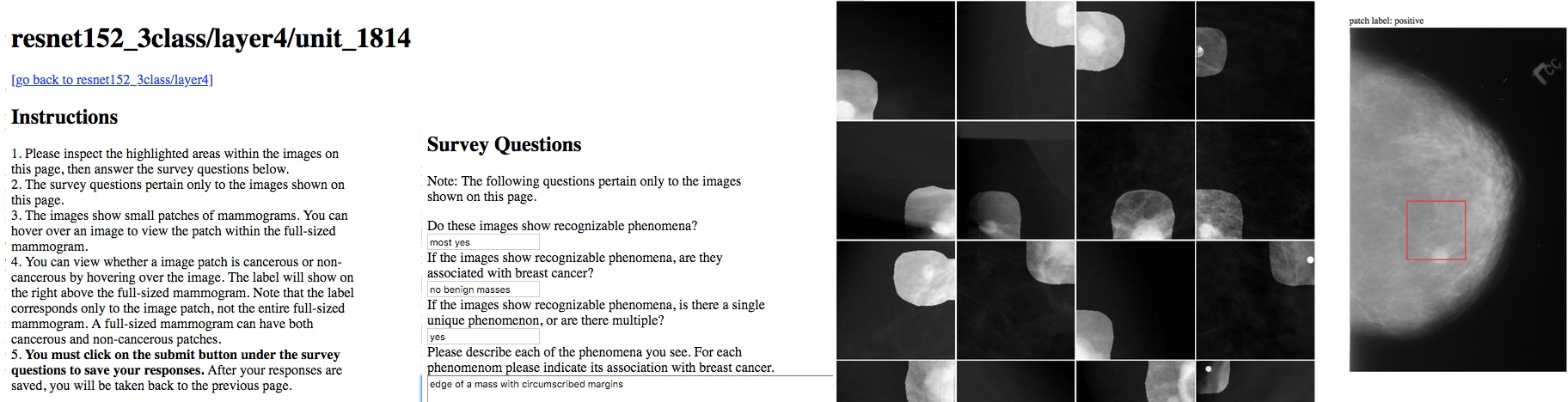}
    \caption{\textit{The interface of the DeepMiner survey:} Experts used this survey form to label influential units. The survey asks questions such as: ``Do these images show recognizable phenomena?'' and ``Please describe each of the phenomena you see. For each phenomenon please indicate its association with breast cancer.'' In the screenshot above, the radiologist who was our expert-in-the-loop has labeled the unit's phenomena as `edge of mass with circumscribed margins'.}
    \label{fig:survey}
\end{figure}

\begin{figure}[ht!]
    \centering 
    \includegraphics[width=0.8\textwidth]{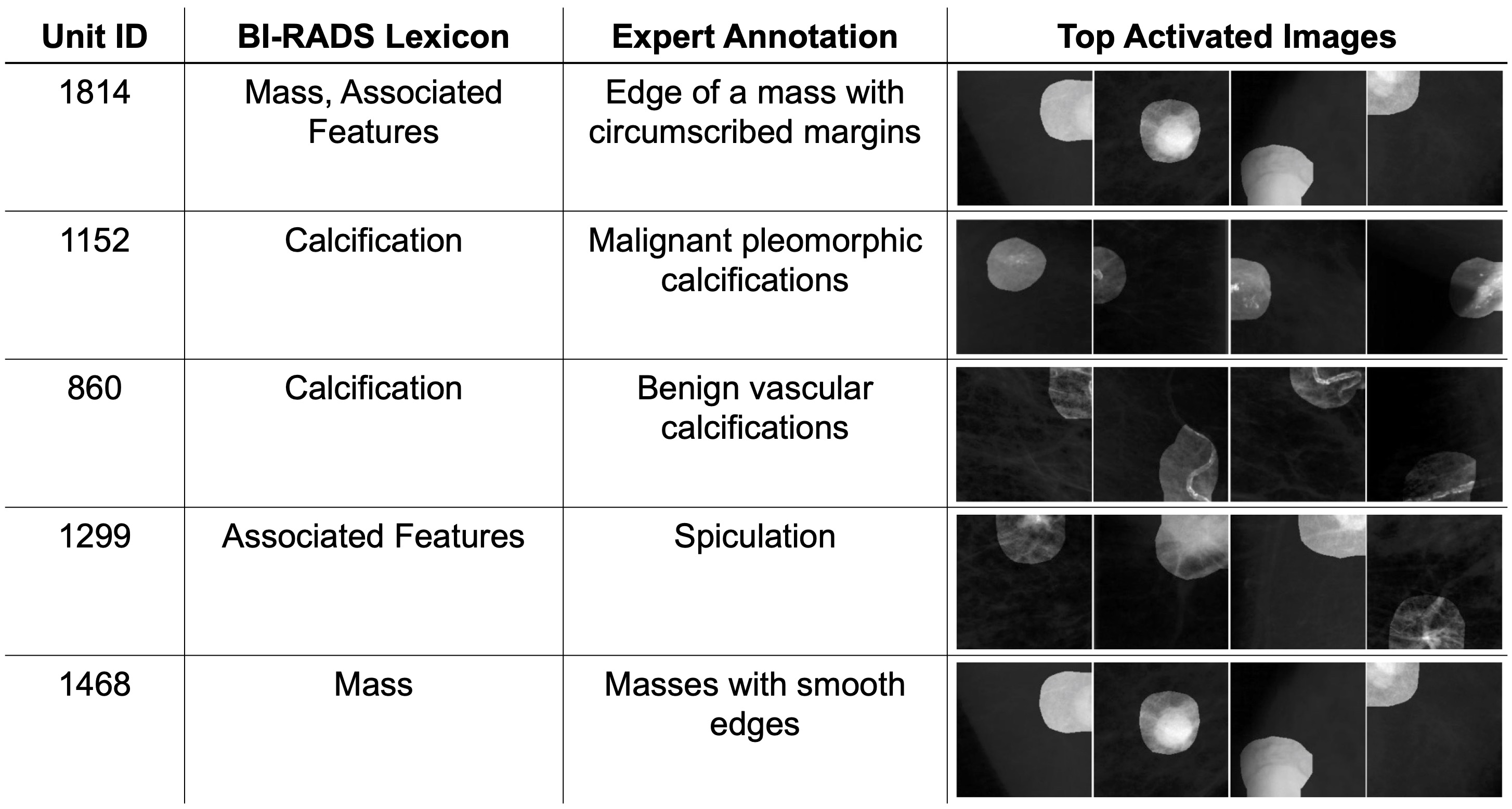}
    \caption{\textit{Interpretable units discovered by DeepMiner:} The table above illustrates five annotated units from the last convolutional layer of our reference network. Even though the CNN presented in this paper was only trained to classify normal, benign, and malignant tissue, these internal units detect a variety of recognizable visual events. Both benign and malignant calcifications are identified, as well as features related to the margins of masses. These details are significant factors in planning interventions for breast cancer. Please refer to the supplement for a full table of annotated units.}
 \label{fig:vis_units}
\end{figure}

We create visualizations for each individual unit by passing every image patch from all mammograms in our hold-out set through our classification network. For each unit in the final convolutional layer, we record the unit's maximum activation value as well as the receptive field from the image patch that caused the measured activation. To visualize each unit (see Figs.~\ref{fig:survey} and~\ref{fig:vis_units}), we display the top activating image patches sorted by their activation score and further segmented by the binarized and upsampled response map of that unit.

A radiologist and a medical physicist specializing in mammography annotated the 60 most frequently influential units we selected. We compare the named phenomena detected by these units to the BI-RADS lexicon~\citep{reporting1998data}.  The experts used the annotation interface shown in Fig.~\ref{fig:survey}. Our survey displays a table of dozens of the top scoring image patches for the unit being visualized. When the expert mouses over a given image patch, the mammogram that the patch came from is displayed on the right with the patch outlined in red. This gives the expert some additional context. From this unit preview, experts are able to formulate an initial hypothesis of what phenomena a unit detects.

Of the 60 units selected, 46 were labeled by at least one expert as detecting a nameable medical phenomenon. Fig.~\ref{fig:vis_units} shows five of the annotated units. In this figure, each row illustrates a different unit. The table lists the unit ID number, the BI-RADS category for the concept the unit is detecting, the expert-provided unit annotation, and a visual representation of the unit. We visualize each unit by displaying the top four activating image patches from the hold-out set. The unit ID number is listed to uniquely identify each labeled unit in the network used in this paper, which will be made publicly available upon publication.

Fig.~\ref{fig:vis_units} demonstrates that the DeepMiner framework discovers significant medical phenomena, relevant to mammogram-based diagnosis. Because breast cancer is a well-characterized disease, we are able to show the extent to which discovered unit detectors overlap with phenomena deemed to be important by the radiological community. For diseases less well understood than breast cancer, DeepMiner could be a useful method for revealing unknown discriminative visual features helpful in diagnosis and treatment planning.

\subsection{Explaining Network Decisions}

We further use the annotated units to build an explanation for single image prediction. We first convert our trained network into a fully convolutional network (FCN) using the method described in \citep{long2015fully} and remove the global average pooling layer. The resulting network is able to take in full mammogram images and output probability maps aligned with the input images.

As illustrated in Fig.~\ref{fig:deepminer}, given an input mammogram, we output a classification as well as the Class Activation Map (CAM) proposed in \citep{zhou2016learning}. We additionally extract the activation maps for the units most influential towards the classification decision. By looking up the corresponding expert annotations for those units, we are able to see which nameable visual phenomena contributed to the network's final classification. For examples of the DeepMiner explanations, please refer to Sec.~\ref{sec:results}.

\subsection{Annotation Efficiency}
Annotating a single unit through the survey shown in \cref{fig:survey} takes approximately 30 seconds. 
As an expert only needs to annotate 60 units per network, this consumes 30 minutes of the expert's time in total.
This level of annotation efficiency is one of the strengths of the DeepMiner framework and should be contrasted with the standard approach to annotation which processes each image in isolation by annotating the image regions with polygons and labeling each region with a medical term.
Typically, annotating a single image would require 1 minute of an expert's time, and annotating the entire DDSM dataset of 10,000 images would therefore consume 10,000 minutes 
(166 hours) 
of expert time.

\subsection{Incorporating Auxiliary Features}
Genetic information is sometimes used to personalize breast cancer screening. Mutations in BRCA1 and BRCA2 are associated with an estimated lifetime risk of breast cancer of approximately 70 percent \citep{KuchenbaeckerJAMA2017}, and patients carrying these mutations are recommended annual contrast-enhanced breast MRI exams in addition to screening mammography as well as the option to undergo prophylactic mastectomy. Polygenic risk scores, which aggregate the risk of several individually less influential gene mutations, are routinely used in the management of detected breast cancer \citep{SparanoNEJM2018}. Similar scores have been proposed for breast cancer screening \citep{mavaddat2019polygenic} but are not yet routinely used.  
When available, these genetic scores can be incorporated into the DeepMiner classifier as auxiliary input features accompanying the mammogram. The interactions between genetic information and the radiologist interpretations can also be exposed using an attention-based mechanism.
A similar approach could be applied to other risk factors, such as smoking status, age, or body mass index.

\newpage

\section{Results}

\subsection{Classifying Mammogram Regions}
\label{sec:res_class}
We benchmark our reference network on the test set patches using the area under the ROC curve (AUC) score. Our network achieves AUCs of 0.862 for the normal class (pAUC @ TPR of 0.8 was 0.142), 0.844 for the benign class (pAUC of 0.136), and 0.872 for the malignant class (pAUC of 0.146).
This performance is comparable to the state-of-the-art AUC score of 0.88~\citep{shen2017end} for single network malignancy on DDSM. For comparison, positive detection rates of human radiologists range from 0.745 to 0.923~\citep{elmore2009variability}. Note that achieving state-of-the-art performance for mammogram classification is not the focus of this work.

\subsection{Explanation for Predictions}
\label{sec:results}
\begin{figure}[t]
    \centering
    \begin{subfigure}[b]{\textwidth}
        \includegraphics[width=\textwidth]{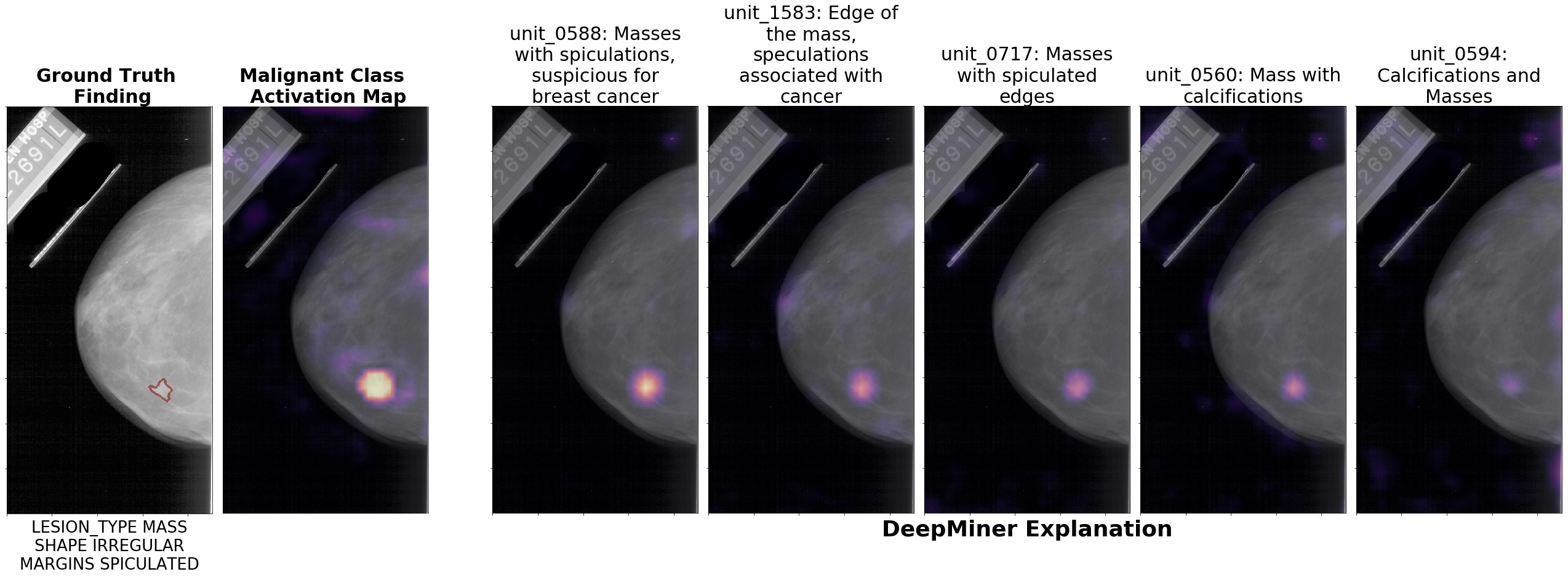}       
	\caption{\footnotesize{The mammogram above is labeled BI-RADS assessment 4 (high risk), DDSM subtlety 2 (not obvious). Our network correctly classifies the mammogram as containing malignancy. Then, DeepMiner shows the most influential units for that classification, which correctly identify the finding as a mass with spiculations.}}
        \label{fig:mtp}
    \end{subfigure}
    \begin{subfigure}[b]{\textwidth} 
        \includegraphics[width=\textwidth]{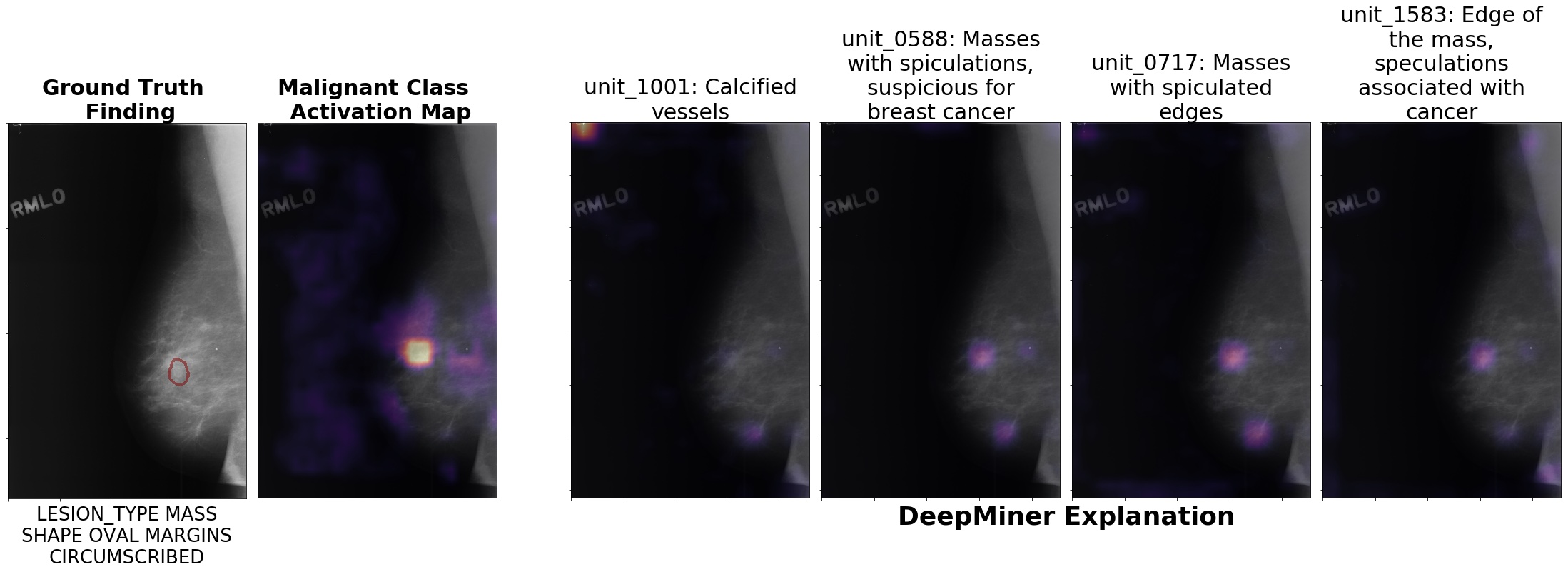}
	\caption{\footnotesize{This mammogram is falsely classified by our network as containing a malignant mass, when it in fact contains a benign mass. However, the DeepMiner explanation lists the most influential unit as detecting calcified vessels, a benign finding, in the same location as the malignant class activation map. The most influential units shown here help explain how the network both identifies a benign event and misclassifies it as a malignant event.}}
	\label{fig:mfp}
    \end{subfigure}  
    \caption{Sample DeepMiner explanations of mammograms classified as malignant. Best viewed in color.}
\label{fig:mexp}
\end{figure}

\begin{figure}[t]
    \centering
    \begin{subfigure}[b]{\textwidth}
        \includegraphics[width=\textwidth]{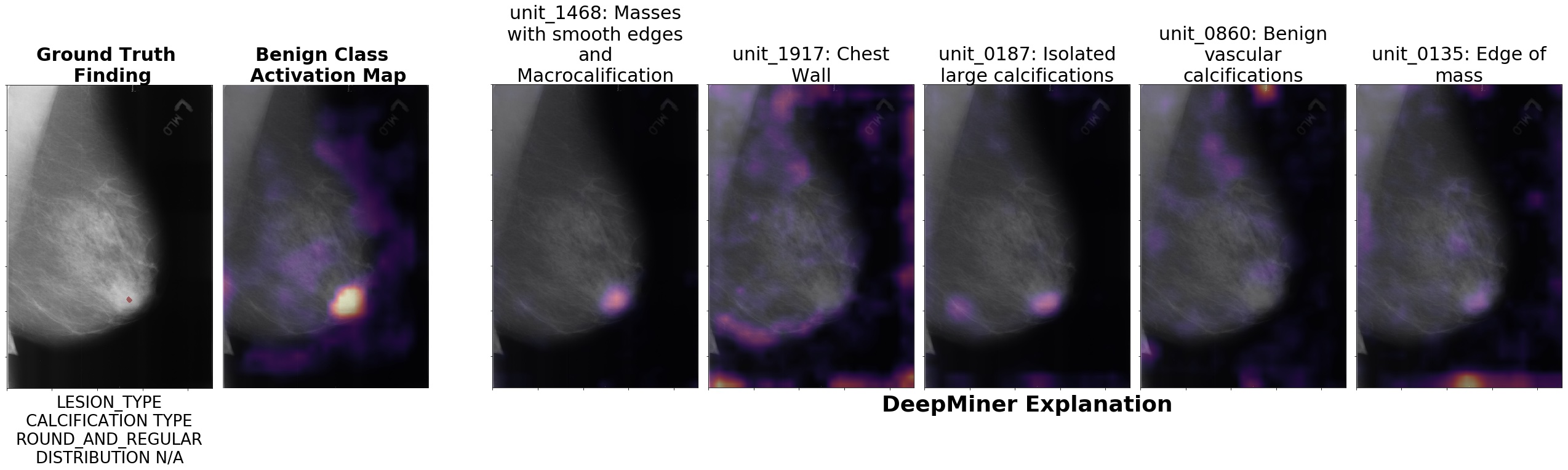}        
	\caption{The above image sequence explains a true positive classification of a benign mammogram. The benign mass is quite small, but several unit detectors identify the location of the true finding as `mass with smooth edges' (likely benign) and `large isolated calcification'.}
        \label{fig:btp}
    \end{subfigure}
    \begin{subfigure}[b]{\textwidth} 
        \includegraphics[width=\textwidth]{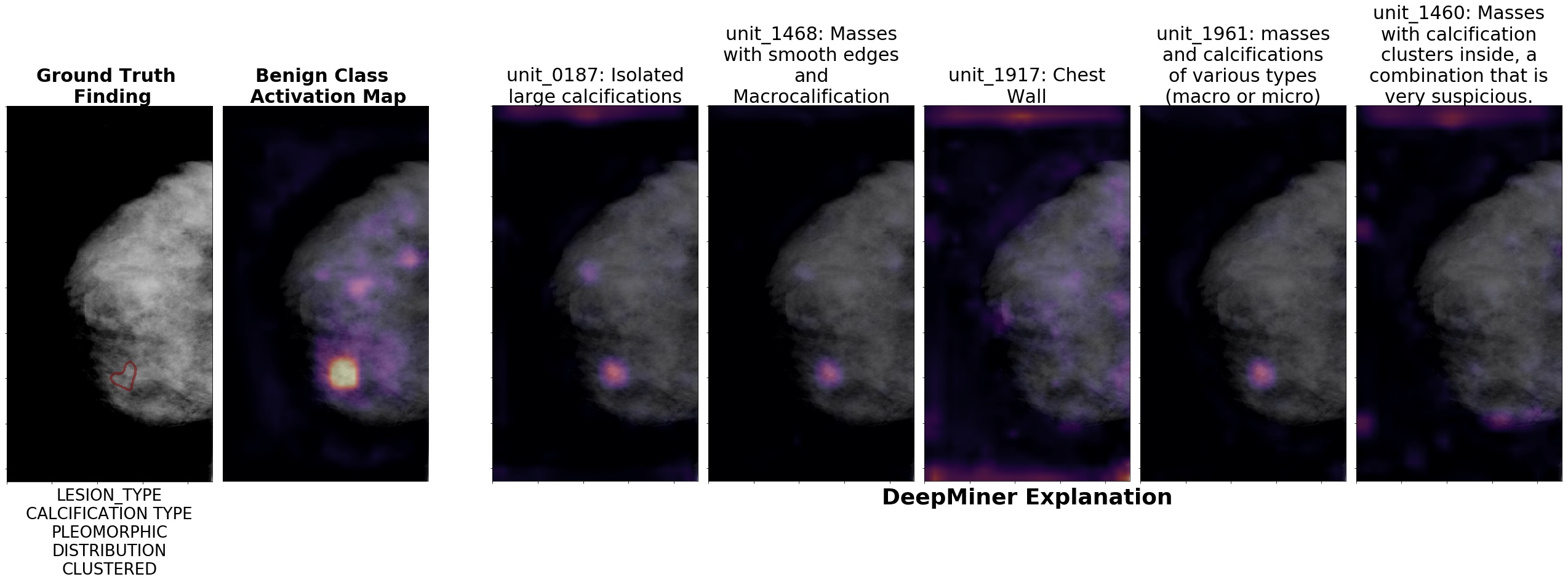}
	\caption{The above image sequence shows a false positive for benign classification. The mammogram actually contains a malignant calcification. However, the 5th most influential unit detected a `mass with calcification clusters inside [...] very suspicious' just below the location of the ground truth finding.}
	\label{fig:bfp}
    \end{subfigure}       
    \caption{Sample DeepMiner explanations of mammograms classified as benign. Best viewed in color.}
\label{fig:bexp}
\end{figure}

Using the DeepMiner framework, we next create explanations for the classifications of our reference network on the test set. Figs.~\ref{fig:mexp} and~\ref{fig:bexp} show sample DeepMiner explanations for malignant and benign classifications, respectively. In these figures, the left-most image is the original mammogram with the benign or malignant lesion outlined in maroon. The ground truth radiologist's report from the DDSM dataset is printed beneath each mammogram. The heatmap directly on the right of the original mammogram is the class activation map for the detected class. 

In Figs.~\ref{fig:mexp} and~\ref{fig:bexp}, the four or five images on the right-hand side show the activation maps of the units most influential to the prediction. In all explanations, the DeepMiner explanation units are among the top eight most influential units overall, but we only print up to five units that have been annotated as part of the explanation.

In these examples, the DeepMiner explanation gives context and depth to the final classification. For the true positive classifications in Figs.~\ref{fig:mtp} and~\ref{fig:btp}, the explanation further describes the finding in a manner consistent with a detailed BI-RADS report. For the false positive cases in Figs.~\ref{fig:mfp} and~\ref{fig:bfp}, the explanation helps to identify why the network is confused or what conflicting evidence there was for the final classification. 

To test whether the DeepMiner explanations enable an expert to better distinguish malignant cases from benign cases, we carried out the following human-in-the-loop experiment.
We selected $165$ benign cases uniformly at random from all benign and benign-without-callback cases in the test set and $165$ malignant cases uniformly at random from all malignant cases in the test set.
For each of the resulting $n=330$ cases, we outputted the expert annotations associated with the three units most influential in the classification decision for that case.
An expert was then tasked with classifying each case as `malignant' or `benign' based only on the knowledge provided by our generated explanations.  
As an example, \cref{table:cancer_vs_benign} displays the first three cases presented to the expert.
Equipped with only the DeepMiner explanations, the expert correctly classified $182$ of the cases, a significant improvement over the baseline of random guessing (the associated p-value was $0.0346$ for the exact test of a binomial proportion exceeding $0.5$). 

\begin{table}[h]
\caption{\label{table:cancer_vs_benign}
DeepMiner explanations presented for the first three cases in the malignant vs.\ benign expert classification experiment of \cref{sec:results}.}
\tiny
\begin{tabular}{cc}
\toprule 
\footnotesize{\textbf{DeepMiner Explanation}} \\  
\midrule
('Calcified vessels', 'Calcified vessels', 'Spiculation')\\
('Calcified vessels', 'Edge of the mass, speculations associated with cancer', 'Malignant pleomorphic calcifications')\\
('Spiculation', 'Calcified vessels', 'Normal Breast Tissue') \\
\bottomrule
\end{tabular}
\end{table}

\section{Conclusion}
We proposed the DeepMiner framework, which uncovers interpretable representations in deep neural networks and builds explanations for deep network predictions. We trained a network for mammogram classification and showed with human expert annotation that interpretable units emerge to detect different types of medical phenomena even though the network is trained using only coarse labels. We further used the expert annotations to automatically build explanations for final network predictions. 
We believe our proposed framework is applicable to many other domains, potentially enabling discovery of previously unknown discriminative visual features relevant to medical diagnosis.

\subsection*{Disclosure Statement}
The authors have no conflicts of interest to declare.

\section*{Acknowledgments}
We thank the editor and anonymous referees for their role in improving this manuscript.

\appendix

\section{An Introduction to Convolutional Neural Networks}
\label{sec:cnn-intro}

Convolutional neural networks (CNNs) are a popular method for extracting high level information from images. In this appendix, we will provide a concise and targeted introduction to CNNs and not to provide a complete reference. Interested readers can consult \cite{Goodfellow-et-al-2016} for more information.

CNNs apply multiple layers of processing to the original image. We will denote each layer as $f_l (x,y,c)$, where $x$ and $y$ denote 2D Cartesian image coordinates, $l$ denotes the layer index, and $c$ denotes the channel index. The first layer, $f_1 (x,y,c)$, is the original image. Most images produced using digital cameras provide three channels for the RGB (red, green, and blue) colors. The grayscale images captured using mammography do not have colors, so the single-channel grayscale is replicated across the three color channels for the first layer. For an $N$-layer CNN, subsequent convolutional layers are defined as follows:
\[ \textstyle f_l (x,y,c)=h(\sum_j f_{l-1} (x,y,j)\circledast n_{lc} (x,y,j) ) \]
The $\circledast$ represents the convolution operator, and the $n_{lc} (x,y,j)$ are the filter weights for the $l^{th}$ layer and $c^{th}$ channel. These filter weights are typically nonzero only in a limited region. Many architectures specify that the filter weights are $3\times3$ or $5\times5$ in size, making the convolution computationally efficient.  Here, $h(x)$ is the nonlinear activation function. A popular choice for $h(x)$ is the rectified linear (ReLU) function, $h(x)=\max(x,0)$. 

In a binary classification task that distinguishes cancer patches from non-cancer patches, to produce an estimate of the probability of cancer, the contents of the final convolutional layer $f_N (x,y,c)$ are rearranged into a single one-dimensional vector, $\tilde{f}_N (t)$. We then compute the cancer probability using the sigmoid function:
\[ p_{\textup{cancer}}=({1+e^{-w^\top \tilde{f}_N} })^{-1} \]
Here, $w$ is a weighting vector and $w^\top$ is its transpose so that $w^\top \tilde{f}_N$ is the dot product between $w$ and $\tilde{f}_N$. 

We have described a simple CNN that only uses convolutional layers followed by a simple fully connected layer to calculate the cancer probability. Most CNNs used today also include more than two output classes (e.g., malignant vs. benign vs. normal) and other kinds of layers, including downsampling layers that reduce the size of the image, allowing future layers to efficiently extract information across larger spatial scales. Skip connections, which connect early layers with later layers, are also widely used, allowing efficient training of deep networks. Deep networks have found much more empirical success than shallow networks in many domains. The ResNet architecture used in this work is best known for introducing skip connections to the CNN literature.

CNNs are typically trained by presenting the network with thousands to millions of labeled datapoints. In the context of our binary classification example, each datapoint is an image patch of a mammogram that is labeled to be either positive (cancer) or negative (not cancer). For each patch, both the filter weights $n_{lc} (x,y,j)$ and the final weighting factor $w$ are updated so that the final probability $p_{cancer}$ is increased for cancerous patches and decreased for negative patches. 

Several algorithms are available to perform the training. Most of the algorithms are a variation on gradient descent, also called the Newton-Raphson method. Stochastic gradient descent (SGD) is one of the most popular choices and uses a limited number of datapoints selected uniformly at random from the full dataset. The number of samples used for each iteration of SGD is known as the minibatch size. The use of a minibatch rather than the entire collection of samples can be viewed as a type of regularization, to prevent the network from collapsing into a local minimum. The step size in SGD must be empirically selected according to a preset schedule. Besides SGD, other optimization algorithms are available which automatically select an adaptive step size, including AdaGrad (adaptive gradient descent) or Adam (adaptive momentum). These often reduce training time, but the selection of the best optimizer for a given dataset remains empirical. 

CNNs often have millions of unknown parameters that must be learned. As the CNN is trained, the filter weights $n_{lc}$ and the final weighting vector $w$ are progressively improved.

\printbibliography
\end{document}

\title{Supplemental Submission for DeepMiner: Discovering Visual Concepts for Mammogram Classification 
and Explanation}
\author{}
\institute{}
\pagestyle{plain}

\maketitle
\begin{figure}[t]
    \centering
    \begin{subfigure}[b]{\textwidth}
        \includegraphics[width=\textwidth]{figures/resnet152_3class_pos-class2_cancer_03-A_1060_1_LEFT_CC.jpg}        
	\caption{This is a true positive malignant classification.}
        \label{fig:}
    \end{subfigure}
    \begin{subfigure}[b]{\textwidth}
        \includegraphics[width=\textwidth]{figures/resnet152_3class_pos-class2_cancer_04-A_1098_1_RIGHT_MLO.jpg}        
	\caption{This is a true positive malignant classification.}
        \label{fig:}
    \end{subfigure}
    \begin{subfigure}[b]{\textwidth}
        \includegraphics[width=\textwidth]{figures/resnet152_3class_cancer_03-A_1060_1_LEFT_CC.jpg}        
	\caption{This is a true positive malignant classification.}
        \label{fig:}
    \end{subfigure}
    \begin{subfigure}[b]{\textwidth} 
        \includegraphics[width=\textwidth]{figures/resnet152_3class_falsepos-class2_benign_06-C_0390_1_RIGHT_MLO.jpg}
	\caption{This is a false positive malignant classification. This mammogram contained a breast labeled 'Pathology Benign' in the DDSM dataset.}
	\label{fig:}
    \end{subfigure}       
    \caption{DeepMiner Explanations of Malignant Classifications}
\label{fig:class2_exp}
\end{figure}

\begin{figure}[b]
    \centering
    \begin{subfigure}[b]{\textwidth}
        \includegraphics[width=\textwidth]{figures/resnet152_3class_pos-class1_a4_s5_benign_08-A_1778_1_LEFT_CC.jpg}        
	\caption{This is a true positive benign classification.}
        \label{fig:}
    \end{subfigure}
    \begin{subfigure}[b]{\textwidth}
        \includegraphics[width=\textwidth]{figures/resnet152_3class_pos-class1_benign_without_callback_01-B_3241_1_RIGHT_CC.jpg}        
	\caption{This is a true positive benign classification.}
        \label{fig:}
    \end{subfigure}
    \begin{subfigure}[b]{\textwidth}
        \includegraphics[width=\textwidth]{figures/resnet152_3class_pos-class1_a0_s2_benign_09-D_4023_1_RIGHT_CC.jpg}        
	\caption{This is a true positive benign classification. The subtlety rating for this finding was a 2 ( on a scale of 0 subtle to 5 obvious).}
        \label{fig:}
    \end{subfigure}
    \begin{subfigure}[b]{\textwidth} 
        \includegraphics[width=\textwidth]{figures/resnet152_3class_false-pos-class1_normal_03-A_0341_1_LEFT_CC.jpg}
	\caption{This is a false positive benign classification. This breast was labeled Normal in the DDSM dataset.}
	\label{fig:}
    \end{subfigure}       
    \caption{DeepMiner Explanations of Benign Classifications}
\label{fig:class1_exp}
\end{figure}

\begin{table}
  \rowcolors{2}{gray!25}{white}
\begin{tabular}{ l  r}
\rowcolor{gray!50}
\textbf{Unit ID} & \textbf{Expert Annotation}\\
  \hline
unit 2038 & Normal Breast Tissue\\
unit 1468 & Masses with smooth edges and Macrocalification\\
unit 1583 & Edge of the mass, speculations associated with cancer\\
unit 0594 & Calcifications and Masses\\
unit 0820 & Mass\\
unit 0281 & arterial calcifications\\
unit 0725 & suspicious calcifications\\
unit 0404 & Normal Breast Tissue\\
unit 1311 & Normal Breast Tissue\\
unit 1299 & Spiculation\\
unit 0560 & Mass with calcifications\\
unit 1373 & Masses with calcifications\\
unit 1168 & Normal Breast Tissue\\
unit 1001 & Calcified vessels\\
unit 1405 & Benign calcifications, vascular clarification\\
unit 1430 & Calcifications and arterial calcification\\
unit 1961 & masses and calcifications of various types (macro or micro)\\
unit 1962 & Calcifications\\
unit 0003 & Microcalcifications, Dense Tissue\\
unit 1549 & Normal Breast Tissue\\
unit 1780 & Edge of breast or with the chest wall\\
unit 1680 & Calcified vessels\\
unit 1152 & Malignant pleomorphic calcifications\\
unit 1118 & Normal Breast Tissue\\
unit 0187 & Isolated large calcifications\\
unit 0588 & Masses with spiculations, suspicious for breast cancer\\
unit 1460 & Masses with calcification clusters inside, a combination that is very suspicious.\\
unit 1917 & Chest Wall\\
unit 1814 & Edge of a mass with circumscribed margins\\
unit 0175 & Normal Breast Tissue\\
unit 0135 & Edge of mass\\
unit 0031 & Normal Breast Tissue\\
unit 0889 & Mass\\
unit 0036 & Mass with circumscribed margins\\
unit 1122 & Pectoralis or inframammary folds\\
unit 1065 & Interface of tissue with air or fat or normal tissue\\
unit 0097 & Normal Breast Tissue\\
unit 1024 & Normal Breast Tissue\\
unit 0091 & edge of a suspicious mass, mass irregularly-shaped\\
unit 0842 & Normal Breast Tissue\\
unit 0860 & Benign vascular calcifications\\
unit 1593 & Normal Breast Tissue\\
unit 1597 & Calcified vessels\\
unit 0717 & Masses with spiculated edges\\
unit 0659 & Calcifications and Masses\\
unit 0344 & Dense Tissue\\
  \hline  
\end{tabular}
\caption{All Annotated Units discovered using the DeepMiner framework from Layer 4 of our reference ResNet152 Network. Many of these units identify benign and malignant phenomena, but a significant number also identify `Normal Tissue'. These units were among those which had high influence scores for the Normal class.}
\end{table}